\documentclass[nohyperref]{article}
\usepackage{microtype}
\usepackage{graphicx}
\usepackage{algorithm}
\usepackage{subfigure}
\usepackage{booktabs} 
\usepackage{hyperref}
\usepackage[accepted]{icml2022}
\usepackage{amsmath}
\usepackage{amssymb}
\usepackage{mathtools}
\usepackage{float}
\usepackage{amsthm}
\usepackage[capitalize,noabbrev]{cleveref}
\theoremstyle{plain}

\theoremstyle{definition}

\theoremstyle{remark}

\usepackage[textsize=tiny]{todonotes}
\icmltitlerunning{FedControl: When Control Theory Meets Federated Learning}

\newcommand{\rtot}[1]{\overline{#1}}

\begin{document}
\twocolumn[
\icmltitle{FedControl: When Control Theory Meets Federated Learning}
\icmlsetsymbol{equal}{*}
\begin{icmlauthorlist}
\icmlauthor{Adnan Ben Mansour}{equal,yyy}
\icmlauthor{Gaia Carenini}{equal,yyy}
\icmlauthor{Alexandre Duplessis}{equal,yyy}
\icmlauthor{David Naccache}{equal,yyy}
\end{icmlauthorlist}
\icmlaffiliation{yyy}{Department of Computer Science,
   ENS, CNRS, PSL Research University,
   Paris, France }
\icmlcorrespondingauthor{Adnan Ben Mansour}{adnan.ben.mansour@ens.psl.eu}
\icmlkeywords{Fedetared Learning, Control Theory, Aggregation}
\vskip 0.3in
]
\printAffiliationsAndNotice{\icmlEqualContribution}
\begin{abstract}
To date, the most popular federated learning algorithms use coordinate-wise averaging of the model parameters. We depart from this approach by differentiating client contributions according to the performance of local learning and its evolution. The technique is inspired from control theory and its classification performance is evaluated extensively in IID framework and compared with FedAvg.
\end{abstract}
\section{Introduction}
Until recently, machine learning models were extensively trained in a data center setting using powerful computing nodes, fast inter-node communication links, and large centrally available training datasets. The future of machine learning lies in moving both data collection as well as model training to the edge. 
Moreover, nowadays, more and more collected data-sets are privacy sensitive, in particular in confidentiality-critical fields such as patients' medical information processing. Specifically, reducing human exposure to data is highly desirable to avoid confidentiality violations due to human failure. This may preclude logging to a data center and performing training there using conventional approaches.  In 2017, \cite{McMahan2017CommunicationEfficientLO} advocated an alternative distributed data-mining technique for edge devices, termed Federated Learning (FL), allowing a decoupling of model training from the need for direct access to the raw training data.\vspace{0.1cm} \\
\emph{Federated Learning} (FL)  is a protocol acting according to Algorithm \ref{alg:generic_FL}, \cite{Li2020FederatedLC}. FL involves a group of devices named \emph{clients} and a server that coordinates the learning process. Each client has a local training data-set which is never uploaded to the \emph{server}. The goal is to train a global model by aggregating local training results.
Parameters, fixed by server, include: a set $I$ grouping \emph{N} clients, the ratio of clients \emph{C} selected at each round, the number of communication rounds \emph{R} and a number of local epochs \emph{E}. We note $\rtot{r} = rE$ the epoch corresponding to round $r$. The model is defined by its weights: at the end of each epoch $t \in \{ 0, \dots, RE -1 \}$, $w_{t+1}^i$ defines the weights of client $i \in I$. For each communication round $r \in \{ 0, 1, \dots, (R-1) \}$, $w_{\rtot{r}}$ is the global model detained by the server and $w_{\rtot{R}}$ is the final weight.
\begin{algorithm}[tb]
\caption{Generic Federated Learning Algorithm}
\label{alg:generic_FL}
\textbf{Input}: \emph{N}, \emph{C}, \emph{R}, \emph{E} ~~~~~~~~
\textbf{Output}: $w_{\rtot{R}}$
\begin{algorithmic}[1] %[1] enables line numbers
\STATE Initialize $w_{\rtot{0}}$.
\FOR{each round $r\in\{0,1, \dots,(R-1)\}$}
%\STATE $m \leftarrow \max(C \cdot N, 1)$
\STATE $I_r \leftarrow \text{random set of $\max(C \cdot N, 1)$ clients}$
\FOR{each client $i \in I_r$ \textbf{in parallel}}
\STATE $w_{\rtot{r+1}}^{i} \leftarrow \text{\textsc{Client-Update}}(w_{\rtot{r}})$
\ENDFOR
\STATE $w_{\rtot{r+1}} \leftarrow \text{\textsc{Aggregation}}(w_{\rtot{r+1}}^1, \dots, w_{\rtot{r+1}}^N)$
\ENDFOR
\STATE \textbf{return} $w_{\rtot{R}}$
\end{algorithmic}
\end{algorithm}
 Algorithm \ref{alg:generic_FL} encodes the training procedure described below. There is a fixed set of $I = \{1,\dots,N\}$ clients (each with a local data-set). Before every communication round $r$ the server randomly selects a set $I_r$ of $C\cdot N$ clients, sends them the current global algorithm state, and then asks them to perform local computations based on the global state and their local data-set, and send back an update. At the end, it updates the weights of the model by aggregating clients' updates and the process repeats. For sake of generality, no structure is specified for the local training procedure (\emph{Client-Update}): several different methods can be employed, for instance mini-batch SGD \cite{Gower2019SGDGA}, Newton methods  or PAGE \cite{DBLP:journals/corr/abs-2108-04755}. Similarly, no function for model aggregation is given yet. 
\paragraph{Model Aggregation} 
To date, several aggregating functions have been proposed to accomplish this task. In 2017,  \cite{McMahan2017CommunicationEfficientLO} proposed a plain coordinate-wise mean averaging of model weights.  \cite{ek:hal-03207411} adjusted this to enforce closeness of local and global updates. \cite{Yurochkin2019BayesianNF} takes the invariance of network weights to permutation into account. More recently, \cite{DBLP:journals/corr/abs-2111-08649} suggested to add a derivative term to FedAvg enhancing the relevance, in the aggregation, of the models of the clients whose performance are rapidly improving. Despite all advances, there is no formal way to establish in the general case which aggregation strategy to select.
\paragraph{Our Contribution} We introduce a new class of aggregation algorithms called FedControl. All class members  contain a differential term and an integral term, augmenting rapidity and stability. Furthermore, we provide an intuitive control theory interpretation of them as PID controllers (cf. \cite{articlePID}). We validate empirically the model and compare it with other state of the art algorithms. 
\section{Structure and Interpretation}
As observed in the concluding remarks of \cite{DBLP:journals/corr/abs-2111-08649},  there is a proper way for interpreting a FL problem as a control problem and then  establish an equivalence between the central server that does the aggregation and a control unit in a feedback loop. Extending the interpretation to the weighted averaging approach, it is natural to interpret the aggregation procedure introduced in \cite{DBLP:journals/corr/abs-2111-08649} as an approximation of a PID controller where the cardinality-dependent and the loss-dependent terms are respectively functioning as the proportional component and the derivative component in a PID controller (PIDC). In the wake of this interpretation, a natural better approximation for PIDCs can be obtained by adding an integral part to FedCostWAvg. From now on, we will refer with the name FedControl to any aggregation algorithm including a proportional component, a derivative component and an integral component.
\subsection{The Class of FedControl methods}
We note $\ell_r^i=F_i(w_{\rtot{r}}^i)$ the local loss of client $i$ at the end of round $r-1$.
The class of FedControl algorithms has an aggregation function of the form: 
\begin{equation}\label{eq_fond}
    w_{\rtot{r}} = \sum\limits_{i=1}^N w_{\rtot{r}}^i\cdot \big(\alpha \frac{s^i}{S} + \beta \frac{d_r^i}{D_r} + (1-\alpha-\beta) \frac{k_r^i}{K_r}\big)
\end{equation}
where $\alpha$ and $\beta$ are positive parameters, $s^i$ is the number of samples in the data-set of client $i$, $d_r^i$ is the \emph{derivative term} associated to client $i$ at round $r$, and $k_r^i$ is the \emph{integral term} associated to client $i$ at round $r$ and 
$S$, $D_r$, $K_r$ are the \emph{normalization terms}, that is:
\begin{equation}
    S=\sum\limits_{i=1}^N s^i,\quad
D_r=\sum\limits_{i=1}^N d_r^i,\quad K_r=\sum\limits_{i=1}^N k_r^i
\end{equation}
\subsubsection{The Derivative Term}
The derivative term is intuitively the component measuring the speed of decaying of the loss. There is neither an evident nor a unique way to encode such a term. For  simplicity, we will study a term inspired by the one introduced in \cite{DBLP:journals/corr/abs-2111-08649}, that is: 
\begin{equation}\label{der_ter}
    d_r^i = \ell_{r-1}^i/\ell_r^i
\end{equation}

\subsubsection{The Integral Term}
The integral term plays instead the role of stabilizer. We have considered the following form for such a term: 
\begin{equation}\label{int_term}
    k_r^i = \sum\limits_{r'=1}^r \lambda^{(r-r')} \ell_{r'}^i
\end{equation}
where $0\leq \lambda \leq 1$ is a weighting parameter.
\section{Experiments}
We studied extensively the behavior of the  aggregation strategies proposed and compared them to \cite{McMahan2017CommunicationEfficientLO}. 
In particular, we evaluate the aggregation strategies' performance in  classification tasks in IID framework in a synthetic distributed setting. 
\subsection{Experimental Set-Up}
\paragraph{Synthetic Datasets} We generate a synthetic dataset corresponding to the IID balanced framework by sorting data according to labels, choosing the cardinality of the different local datasets and distributing the items preserving an identical label distribution over the clients. 
\paragraph{Tasks description}
We measure the performance of our aggregation strategies on a classification task over the dataset Fashion-MNIST \cite{https://doi.org/10.48550/arxiv.1708.07747}. 
\paragraph{Model}
The model used on FMINIST  items is a CNN with two $3\times 3$ convolution layers (the first with 32 channels, the second with 64, each followed with $2\times 2$ max pooling), a fully connected layer with 1600 units and ReLu activation, and a final softmax output layer. The local learning algorithm is a simple mini-batch SGD with a batch size fixed at $64$.
\paragraph{Hyperparameters}
We ran all tasks for a small number of communication rounds (between $100-150$) that is enough to investigate the initial convergence speed and with a total number of clients equal to $100$. The decreasing learning rate $\eta_r$ is set to $10^{-3}\cdot 0.99^r$ for each communication round $r \in \{ 0, \dots R-1 \}$. The parameter choices follow from the standard assumptions (dimension of the batch) and from an extensive empirical selection aimed to maximize the accuracy reached after 100 communication rounds. We fix the values of the coefficients $\alpha, \beta$ to $1/3$.
\paragraph{Evaluation}
In order to evaluate our family of algorithm, we compare the performance of FedControl with the one of FedAvg and FedWCostAvg in IID framework, by measuring both the accuracy on $100$ global epochs, and the $R_{60}$ which is the number of communication rounds required to reach $60\%$ accuracy.\\
\subsection{Resources}
The Pytorch code has been trained on a NVIDIA Tesla V100 GPU, with 16 GB of GPU memory on an AWS instance for $\simeq 50$ hours, followed by $\simeq 50$ hours on a NVIDIA RTX 3090 with $24$ GB of GPU memory.
\subsection{Experimental Results}
\begin{figure}
    \centering
    \includegraphics[width=0.91\linewidth]{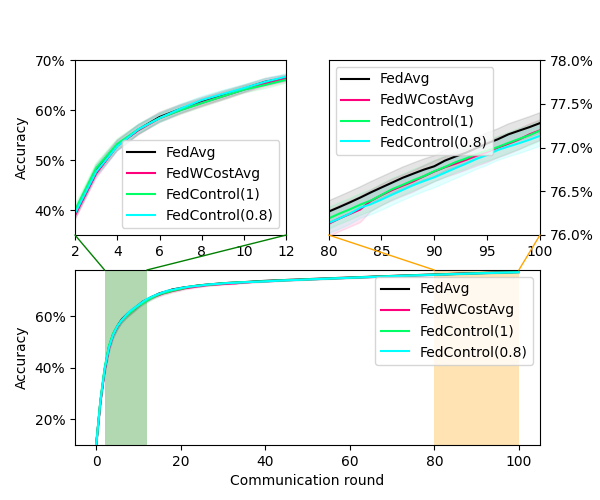}
    \caption{Accuracy over rounds on FMNIST IID with confidence interval at 95\%}
    \label{fig:figure1}
\end{figure}

\begin{figure}
    \centering
    \includegraphics[width=0.91\linewidth]{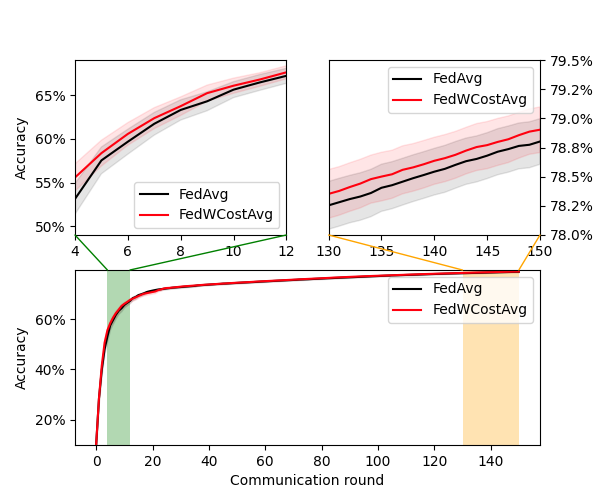}
    \caption{Accuracy over rounds on FMNIST non-IID with confidence interval at 95\%}
    \label{fig:leonVSavg}
\end{figure}

The results of the experiment evidence how the performance of all the methods are almost indistinguishable in terms of accuracy and $R_{60}$ (Figure \ref{fig:figure1}).\\
Only FedAvg seems to perform slightly better than the other methods. This advantage of FedAvg seems related to the experimental context. In fact, in a supplementary experience, conducted in Non-IID framework upon FMNIST with parameters $\eta_r= 10^{-3}\cdot 0.99^r$, where we compare on $150$ global epochs the performance of FedWCostAvg with the one of FedAvg, the results are reversed (Figure \ref{fig:leonVSavg}).\\
Concerning instead the $R_{60}$, we get the results below: 
\begin{center}
\begin{tabular}{ *5c }    \toprule
\emph{Strategies} & \emph{$\textrm{R}_{60}$} \\\midrule
FedAvg  &   $6.198 \pm 0.122$  & \\
FedWCostAvg  &  $6.269 \pm 0.143$ & \\
FedControl($\lambda = 1$) & $6.215 \pm 0.130$ \\
FedControl($\lambda = 0.8$) & $6.375 \pm 0.140$ \\\bottomrule
 \hline
\end{tabular}  
\end{center}
\section{Discussion and Conclusion}
\paragraph{Contribution}
In this paper, we have presented a new class of aggregation strategies designed by analogy with PID controllers. These aggregations establish every round the contribution of each client according to to the performance of local learning in the previous round, and keeping trace of the whole training history. We have tested these methods extensively concluding that their performances are comparable with the ones of the best algorithms at the state of the art. 
\paragraph{Limitation and Further work}
Several limitations are present in this work and may be the starting point for further investigations. We list the main ones below in increasing order of importance.\\
The first point of weakness concerns the lack of tuning for parameters $\alpha$ and $\beta$ that have been fixed to $\frac{1}{3}$ to study the case of balanced influence of the averaging, derivative and integral terms. An optimal choice of these constants could have led to better results (at least on our dataset), since this choice is a token of the trade-off between stability and speed.\\
The second point concerns the tuning for parameter $\lambda$. We did it in a preliminary phase but not extensively enough to obtain results of interest. \\
The third point concerns instead the IID character of the testing framework  that may be the one where our methods perform the worst: in fact FedWCostAvg, which is FedControl(0), seems to perform far better than standard averaging in strongly non-IID environments and in more complex tasks, namely tumor segmentation \cite{DBLP:journals/corr/abs-2111-08649}.\\
The last point is the testing dataset size and the number of reproductions of the experiments, that are both too low and must be improved in more advanced phases of this research. 
\newpage
\bibliography{main}
\bibliographystyle{icml2022}

%%%%%%%%%%%%%%%%%%%%%%%%%%%%%%%%%%%%%%%%%%%%%%%%%%%%%%%%%%%%%%%%%%%%%%%%%%%%%%%
%%%%%%%%%%%%%%%%%%%%%%%%%%%%%%%%%%%%%%%%%%%%%%%%%%%%%%%%%%%%%%%%%%%%%%%%%%%%%%%
% APPENDIX
%%%%%%%%%%%%%%%%%%%%%%%%%%%%%%%%%%%%%%%%%%%%%%%%%%%%%%%%%%%%%%%%%%%%%%%%%%%%%%%
%%%%%%%%%%%%%%%%%%%%%%%%%%%%%%%%%%%%%%%%%%%%%%%%%%%%%%%%%%%%%%%%%%%%%%%%%%
\end{document}